\documentclass[10pt,twocolumn,letterpaper]{article}

\usepackage{cvpr}              % To produce the CAMERA-READY version
\usepackage{graphicx}
\usepackage{amsmath}
\usepackage{amssymb}
\usepackage{booktabs}
\usepackage{color}
\usepackage{alltt}
\usepackage{amsfonts}
\usepackage{mathrsfs}
\usepackage{algorithm}
\usepackage{algorithmic}
\usepackage{multicol,multirow}
\usepackage[dvipsnames]{xcolor,colortbl}
\usepackage{makecell}
\usepackage[accsupp]{axessibility}

\usepackage[pagebackref,breaklinks,colorlinks]{hyperref}

\usepackage[capitalize]{cleveref}
\crefname{section}{Sec.}{Secs.}
\Crefname{section}{Section}{Sections}
\Crefname{table}{Table}{Tables}
\crefname{table}{Tab.}{Tabs.}

\def\cvprPaperID{142} % *** Enter the CVPR Paper ID here
\def\confName{CVPR}
\def\confYear{2022}

\begin{document}

\title{Decoupled Knowledge Distillation}

\author{Borui Zhao$^{1}$ \quad Quan Cui$^{2}$ \quad Renjie Song$^{1}$ \quad Yiyu Qiu$^{1,3}$ \quad  Jiajun Liang$^{1}$ \\
\vspace{-10pt}\\
$^{1}$MEGVII Technology \quad $^{2}$Waseda University \quad $^{3}$Tsinghua University\\
{\tt\small zhaoborui.gm@gmail.com,  cui-quan@toki.waseda.jp,} \\ 
{\tt\small chouyy18@mails.tsinghua.edu.cn,  \{songrenjie, liangjiajun\}@megvii.com} 
}

\maketitle

%%%%%%%%% ABSTRACT
\begin{abstract}
   State-of-the-art distillation methods are mainly based on distilling deep features from intermediate layers, while the significance of logit distillation is greatly overlooked.
   To provide a novel viewpoint to study logit distillation, we reformulate the classical KD loss into two parts, \ie, target class knowledge distillation (TCKD) and non-target class knowledge distillation (NCKD).
   We empirically investigate and prove the effects of the two parts: TCKD transfers knowledge concerning the ``difficulty'' of training samples, while NCKD is the prominent reason why logit distillation works.
   More importantly, we reveal that the classical KD loss is a coupled formulation, which (1) \textbf{suppresses the effectiveness of NCKD} and (2) \textbf{limits the flexibility to balance these two parts}.
   To address these issues, we present Decoupled Knowledge Distillation~(DKD), enabling TCKD and NCKD to play their roles more efficiently and flexibly. Compared with complex feature-based methods, our DKD achieves comparable or even better results and has better training efficiency on CIFAR-100, ImageNet, and MS-COCO datasets for image classification and object detection tasks. This paper proves the great potential of logit distillation, and we hope it will be helpful for future research. 
  The code is available at \href{https://github.com/megvii-research/mdistiller}{https://github.com/megvii-research/mdistiller}.
\end{abstract}

%%%%%%%%% BODY TEXT

%%%%%%%%%%%%%%%%%%%%%
%%%%%%% INTRODUCTION
% 介绍论文的背景，简略说明论文的核心贡献
%%%%%%%%%%%%%%%%%%%%%
\section{Introduction}

\begin{figure}
\setlength{\abovecaptionskip}{5pt} %figurespace
  \centering
  \begin{subfigure}{\linewidth}
%    \fbox{\rule{0pt}{2in} \rule{.9\linewidth}{0pt}}
	\includegraphics[width=\textwidth]{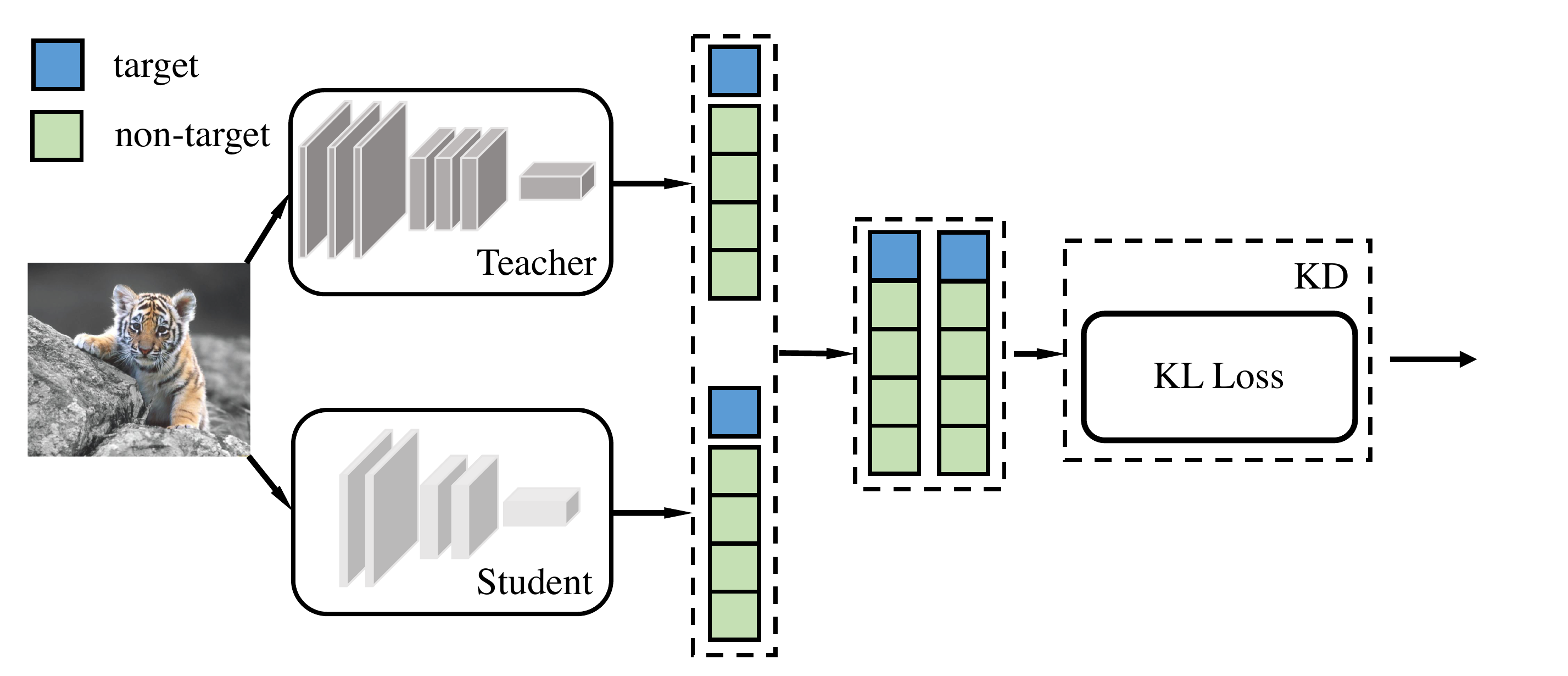}
    \caption{Classical Knowledge Distillation~(KD).}
    \label{fig1a}
  \end{subfigure}
  \begin{subfigure}{\linewidth}
%    \fbox{\rule{0pt}{2in} \rule{.9\linewidth}{0pt}}
	\includegraphics[width=\textwidth]{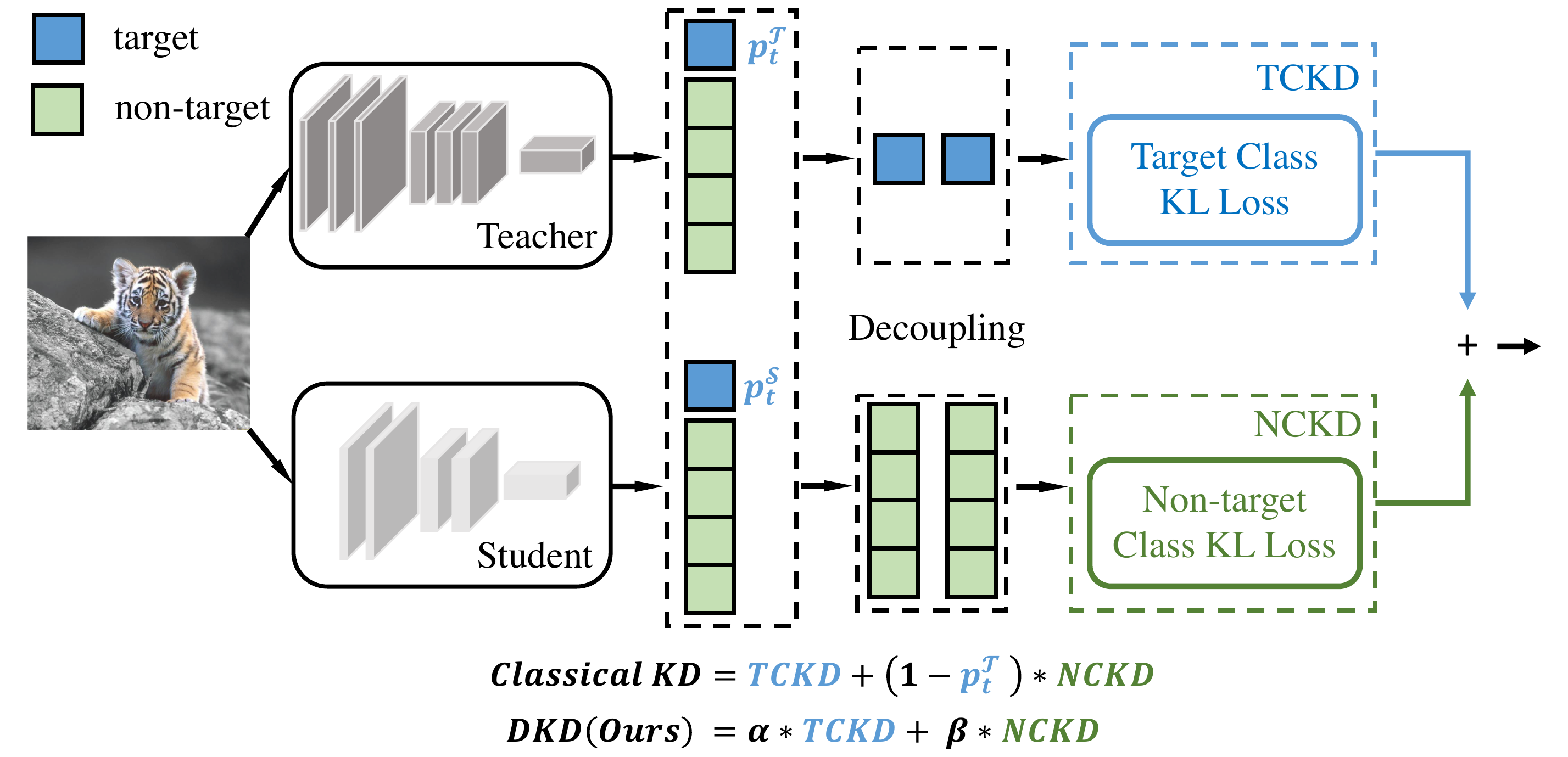}
    \caption{Decoupled Knowledge Distillation~(DKD).}
    \label{fig1b}
  \end{subfigure}
\setlength{\belowcaptionskip}{-15pt} %endfigspace
\caption{Illustration of the classical KD~\cite{kd} and \textbf{our DKD}. 
We reformulate KD into a weighted sum of two parts, \ie, TCKD and NCKD.
The first equation shows that KD (1) couples NCKD with $p_t^{\mathcal{T}}$~(the teacher's confidence on the target class), and (2) couples the importance of two parts. Furthermore, we demonstrate that the first coupling suppresses the effectiveness, and the second limits the flexibility for knowledge transfer. We propose DKD to address these issues, which employs hyper-parameters $\alpha$ for TCKD and $\beta$ for NCKD, killing the two birds with one stone.}
\end{figure}

% 介绍KD的背景
In the last decades, the computer vision field has been revolutionized by deep neural networks~(DNN), which successfully boost various real-scenario tasks, \eg, image classification~\cite{resnet,senet,shufflenetv2}, objection detection~\cite{faster_rcnn,mask}, and semantic segmentation~\cite{shelhamer2016fully,pspnet}. However, powerful networks normally benefit from large model capacities, introducing high computational and storage costs. Such costs are not preferable in industrial applications, where lightweight models are widely deployed. 
In the literature, a potential direction of cutting down the costs is knowledge distillation~(KD). KD represents a series of methods concentrating on transferring knowledge from a heavy model~(teacher) to a light one (student), which can improve the light model's performance without introducing extra costs.

The concept of KD was firstly proposed in~\cite{kd} to transfer the knowledge via minimizing the KL-Divergence between prediction \textit{logits} of teachers and students~(Figure \ref{fig1a}). 
Since~\cite{fitnets}, most of the research attention has been drawn to distill knowledge from deep \textit{features} of intermediate layers. 
Compared with logits-based methods, the performance of feature distillation is superior on various tasks, so research on logit distillation has been barely touched.
% However, training costs of feature-based methods are unsatisfactory, because the essential complex operations and network modules introduce extra computational and storage usage during training time. 
However, training costs of feature-based methods are unsatisfactory, because extra computational and storage usage are introduced~(\eg, network modules and complex operations) for distilling deep features during training time.

% 这篇论文我们干了什么
% 一定程度上拉踩 
% Training costs of logit distillation are not as high as those of feature distillation.
Logit distillation requires marginal computational and storage costs, but the performance is inferior.
Intuitively, logit distillation should achieve comparable performance as feature distillation, since logits are in higher semantic level than deep features. We suppose that the potential of logit distillation is limited by unknown reasons, causing the unsatisfactory performance. 
To revitalize logits-based methods, we start this work by delving into the mechanism of KD.
Firstly, we divide a classification prediction into two levels: (1) a binary prediction for the target class and \textit{all the non-target classes} and (2) a multi-category prediction for \textit{each non-target class}.  %为什么要拆分成target和non-target两部分：把和CE loss重叠的部分给解耦出来，具体来说，CE更加关注的是target class上的准确性
Based on this, we reformulate the classical KD loss~\cite{kd} into two parts, as shown in Figure~\ref{fig1b}. 
One is a binary logit distillation for the target class and the other is a multi-category logit distillation for non-target classes.  
For simplification, we respectively name them as target classification knowledge distillation~(TCKD) and non-target classification knowledge distillation~(NCKD). The reformulation allows us to study the effects of the two parts independently.

% 数学形式完成拆分以后，我们分别探究了两部分的作用，
TCKD transfers knowledge via binary logit distillation, which means only the prediction of the target class is provided while the specific prediction of each non-target class is unknown. A reasonable hypothesis is that TCKD transfers knowledge about the ``difficulty'' of training samples, \ie, the knowledge describes how difficult it is to recognize each training sample.
To validate this, we design experiments from three aspects to increase the ``difficulty'' of training data, \ie, stronger augmentation, noisier label and inherently challenging dataset. 

NCKD only considers the knowledge among non-target logits. 
Interestingly, we empirically prove that only applying NCKD achieves comparable or even better results than the classical KD, indicating the vital importance of knowledge contained in non-target logits, which could be the prominent ``dark knowledge''.

% 接下来，我们根据拆分后的表达式，发现了KD的一些问题
More importantly, our reformulation demonstrates that the classical KD loss is a highly coupled formulation~(as shown in Figure~\ref{fig1b}), which could be the reason why the potential of logit distillation is limited. 
Firstly, the NCKD loss term is weighted by a coefficient that negatively correlates with the teacher's prediction confidence on the target class. Thus larger prediction scores would lead to smaller weights. The coupling significantly suppresses the effects of NCKD on well-predicted training samples.
Such suppression is not preferable since \textit{the more confident the teacher is in the training sample, the more reliable and valuable knowledge it could provide}.
Secondly, the significance of TCKD and NCKD are coupled, \ie, weighting TCKD and NCKD separately is not allowed.
Such limitation is not preferable since \textit{TCKD and NCKD should be separately considered since their contributions are from different aspects}.%due to the inherent difference between the two types of knowledge

% 提出DKD，改进了两个问题
To address these issues, we propose a flexible and efficient logit distillation method named Decoupled Knowledge Distillation~(DKD, Figure \ref{fig1b}). DKD decouples the NCKD loss from the coefficient negatively correlated with the teacher's confidence by replacing it with a constant value, improving the distillation effectiveness on well-predicted samples. Meanwhile, NCKD and TCKD are also decoupled so that their importance can be separately considered by adjusting the weight of each part.

% 总结本文核心贡献
Overall, our contributions are summarized as follows:
\begin{itemize}
\vspace{-10pt}
\setlength{\itemsep}{0pt}
\setlength{\parsep}{0pt}
\setlength{\parskip}{0pt}
% 提供了一个新的分析KD的角度，KD拆成了两部分，分别被分析了
\item We provide an insightful view to study logit distillation by dividing the classical KD into TCKD and NCKD. Additionally, the effects of both parts are respectively analyzed and proved. 
% 根据新的拆分表达式，发现了KD的coupled formulation带来的两个问题
\item We reveal limitations of the classical KD loss caused by its highly coupled formulation. Coupling NCKD with the teacher's confidence suppresses the effectiveness of knowledge transfer. Coupling TCKD with NCKD limits the flexibility to balance the two parts.
% 提出方法+效果很好+高效蒸馏
\item We propose an effective logit distillation method named DKD to overcome these limitations. DKD achieves state-of-the-art performances on various tasks. We also empirically validate the higher training efficiency and better feature transferability of DKD compared with feature-based distillation methods.
\end{itemize}

%%%%%%%%%%%%%%%%%%%%%
%%%%%%% RELATED WORK
% 描述一些相关工作
%%%%%%%%%%%%%%%%%%%%%
\section{Related work}
The concept of knowledge distillation~(KD) was firstly proposed by Hinton \etal in~\cite{kd}. KD defines a learning manner where a bigger teacher network is employed to guide the training of a smaller student network for many tasks~\cite{kd,detmimick1,lizheng}. The ``dark knowledge'' is transferred to students via soft labels from teachers. For raising the attention on negative logits, the hyper-parameter temperature was introduced. 
The following works can be divided into two types, distillation from logits~\cite{eskd,ban,takd,snapshot,mutual} and intermediate features~\cite{ofd,ab,nst,ft,rkd,cc,fitnets,crd,sp,fsp,at}. 

Previous works of logit distillation mainly focus on proposing effective regularization and optimization methods rather than novel methods. DML~\cite{mutual} proposes a mutual learning manner to train students and teachers simultaneously. TAKD~\cite{takd} introduces an intermediate-sized network named ``teacher assistant'' to bridge the gap between teachers and students. Besides, several works also focus on interpreting the classical KD method\cite{quant_kd,understandingkd1}. 

State-of-the-art methods are mainly based on intermediate features, which can directly transfer representations from the teacher to the student~\cite{ofd,ab,fitnets} or transfer the correlation between samples captured in the teacher to the student~\cite{rkd,crd,sp}. 
Most of the feature-based methods could achieve preferable performances~(significant higher than logits-based methods), yet involving considerably high computational and storage costs.

% In this paper, we focus on analyzing what limits the potential of logits-based methods and revitalizing logit distillation.
This paper focuses on analyzing what limits the potential of logits-based methods and revitalizing logit distillation.

%%%%%%%%%%%%%%%%%%%%%
%%%%%%% RETHINKING
% 分析的核心部分，包括提出dkd
%%%%%%%%%%%%%%%%%%%%%
\section{Rethinking Knowledge Distillation}

In this section, we delve into the mechanism of knowledge distillation. We reformulate KD loss into a weighted sum of two parts, one is relevant to the target class, and the other is not. We explore the effect of each part in the knowledge distillation framework and reveal some limitations of the classical KD. Inspired by the findings, we further propose a novel logit distillation method, achieving remarkable performance on various tasks.

% 重写KD表达式
\subsection{Reformulating KD}
\label{sec:reform_kd}

\vspace{5pt}
\noindent \textbf{Notations.}
For a training sample from the $t$-th class, the classification probabilities can be denoted as $\mathbf{p}=[p_1, p_2, ..., p_{t}, ..., p_{C}]\in \mathbb{R}^{1\times C}$, where $p_{i}$ is the probability of the $i$-th class and $C$ is the number of classes. Each element in $\mathbf{p}$ can be obtained by the softmax function:
\begin{equation}
    p_{i} = \frac{\exp(z_{i})}{\sum_{j=1}^{C} \exp(z_{j})},
\label{defpi}
\end{equation}
where $z_i$ represents the logit of the $i$-th class.

To separate the predictions relevant and irrelevant to the target class, we define the following notations. $\mathbf{b}=[p_t, p_{\backslash t}]\in \mathbb{R}^{1\times 2}$ represents the \textit{binary probabilities} of the target class~($p_t$) and all the other non-target classes~($p_{\backslash t}$), which can be calculated by:
\begin{equation*}
	p_{t} = \frac{\exp(z_{t})}{\sum_{j=1}^{C} \exp(z_{j})}, p_{\backslash t} = \frac{\sum_{k=1,k\neq t}^{C} \exp(z_{k})}{\sum_{j=1}^{C} \exp(z_{j})}.
\end{equation*}
Meanwhile, we declare
$\hat{\mathbf{p}}=[\hat{p}_{1}, ..., \hat{p}_{t-1}, \hat{p}_{t+1}, ..., \hat{p}_{C}] \in \mathbb{R}^{1\times (C-1)}$ to independently model probabilities among non-target classes~(\ie, without considering the $t$-th class). Each element is calculated by:
\begin{equation}
    \hat{p}_{i} = \frac{\exp(z_{i})}{\sum_{j=1,j\neq t}^{C} \exp(z_{j})}.
\label{defP}
\end{equation}

\vspace{5pt}
\noindent \textbf{Reformulation.}
In this part\footnote{More mathematical formulations are in the supplement.}, we attempt to reformulate KD with the binary probabilities $\mathbf{b}$ and the probabilities among non-target classes $\hat{\mathbf{p}}$. $\mathcal{T}$ and $\mathcal{S}$ denote the teacher and the student, respectively.
The classical KD uses KL-Divergence as the loss function, which can be written as\footnote{We omit the temperature~(T) in \cite{kd} without loss of generality}:
\vspace{-1pt}
\begin{equation}
    \begin{split}
    \text{KD} &= \text{KL}(\mathbf{p}^{\mathcal{T}}|| \mathbf{p}^{\mathcal{S}})\\
    &=p^{\mathcal{T}}_{t}\log(\frac{p^{\mathcal{T}}_{t}}{p^{\mathcal{S}}_{t}}) + \sum_{i=1,i\neq t}^{C} p^{\mathcal{T}}_{i}\log(\frac{p^{\mathcal{T}}_{i}}{p^{\mathcal{S}}_{i}}).
    \end{split}
    \label{kd_ori_form}
\end{equation}
According to Eqn.(\ref{defpi}) and Eqn.(\ref{defP}) we have $\hat{p}_{i}=p_{i}/p_{\backslash t}$, so we can rewrite Eqn.(\ref{kd_ori_form}) as:
\vspace{-1pt}
\begin{small}
\begin{equation}
	\begin{split}
	\text{KD} &= p^{\mathcal{T}}_{t}\log(\frac{p^{\mathcal{T}}_{t}}{p^{\mathcal{S}}_{t}}) + p^{\mathcal{T}}_{\backslash t}\sum_{i=1,i\neq t}^{C} \hat p^{\mathcal{T}}_{i}(\log(\frac{\hat p^{\mathcal{T}}_{i}}{\hat p^{\mathcal{S}}_{i}}) +\log(\frac{p^{\mathcal{T}}_{\backslash t}}{p^{\mathcal{S}}_{\backslash t}})) \\
	&= \underbrace{p^{\mathcal{T}}_{t}\log(\frac{p^{\mathcal{T}}_{t}}{p^{\mathcal{S}}_{t}}) + p^{\mathcal{T}}_{\backslash t} \log(\frac{p^{\mathcal{T}}_{\backslash t}}{p^{\mathcal{S}}_{\backslash t}})}_{{\text{KL}(\mathbf{b}^{\mathcal{T}}||\mathbf{b}^{\mathcal{S}})}} + p^{\mathcal{T}}_{\backslash t} \underbrace{\sum_{i=1,i\neq t}^{C} \hat p^{\mathcal{T}}_{i} \log(\frac{\hat p^{\mathcal{T}}_{i}}{\hat p^{\mathcal{S}}_{i}})}_{\text{KL}(\hat{\mathbf{p}}^{\mathcal{T}} || \hat{\mathbf{p}}^{\mathcal{S}})}.
	\end{split}
	\label{reform_kd1}
\end{equation}
\end{small}
Then, Eqn.(\ref{reform_kd1}) can be rewritten as:
\begin{equation}
\begin{split}
\text{KD} &= \text{KL}(\mathbf{b}^{\mathcal{T}}||\mathbf{b}^{\mathcal{S}}) + (1-p_{t}^{\mathcal{T}}) \text{KL}(\hat{\mathbf{p}}^{\mathcal{T}} || \hat{\mathbf{p}}^{\mathcal{S}})
\end{split}
\label{form_dkd}
\end{equation}

%KD可以分成两部分，分别叫TCKD和NCKD.接下来看不同作用
As reflected by Eqn.(\ref{form_dkd}), the KD loss is reformulated into a weighted sum of two terms. $\text{KL}(\mathbf{b}^{\mathcal{T}}||\mathbf{b}^{\mathcal{S}})$ represents the similarity between the teacher's and student's binary probabilities of the target class. Thus, we name it Target Class Knowledge Distillation(TCKD). 
Meanwhile, $\text{KL}(\hat{\mathbf{p}}^{\mathcal{T}} || \hat{\mathbf{p}}^{\mathcal{S}})$ represents the similarity between the teacher's and student's probabilities among non-target classes, named Non-Target Class Knowledge Distillation(NCKD). Eqn.(\ref{form_dkd}) could be rewritten as:
\begin{equation}
\begin{split}
\text{KD} &= \text{TCKD} + (1-p_{t}^{\mathcal{T}}) \text{NCKD}.
\end{split}
\label{form_general}
\end{equation}
Obviously, the weight of NCKD is coupled with $p_{t}^{\mathcal{T}}$. 

The reformulation above inspires us to investigate the \textit{individual} effects of TCKD and NCKD, which will reveal the limitations of the classical coupled formulation.

%KD两部分的作用
\subsection{Effects of TCKD and NCKD}
\label{sec:findings}

\vspace{5pt}
\noindent \textbf{Performance gain of each part.}
We individually study the effects of TCKD and NCKD on CIFAR-100\cite{cifar}. 
% Specifically, we explore the distillation performance with either TCKD or NCKD.
ResNet\cite{resnet}, WideResNet (WRN)\cite{wrn} and ShuffleNet\cite{shufflenetv2} are selected as training models, among which both the same and different architectures are considered.
The experimental results are reported in Table~\ref{tab:tckd_nckd}. 
For each teacher-student pair, we report the results of (1) the student baseline~(vanilla training), (2) the classical KD~(where TCKD and NCKD are both used), (3) singly TCKD and (4) singly NCKD. The weight of each loss is set as 1.0~(including the default cross-entropy loss). Other implementation details are the same as those in Sec~\ref{sec:exp}.

\begin{table}[h]
\setlength{\belowcaptionskip}{-10pt} %endtabspace
\center
\begin{small}
\begin{tabular}{ccccc}
\multicolumn{1}{c|}{student}                 & TCKD & \multicolumn{1}{c|}{NCKD} & top-1 & $\Delta$ \\ \Xhline{3\arrayrulewidth}
\multicolumn{5}{c}{\textit{ResNet32$\times$4 as the teacher}}                                                         \\ \hline
\multicolumn{1}{c|}{\multirow{4}{*}{ResNet8$\times$4}} & \space    & \multicolumn{1}{c|}{\space}    & 72.50     & -    \\
\multicolumn{1}{c|}{}                        & \checkmark    & \multicolumn{1}{c|}{\checkmark}    & 73.63     & \textcolor{ForestGreen}{+1.13}    \\
\multicolumn{1}{c|}{}                        & \checkmark    & \multicolumn{1}{c|}{\space}    & 68.63     & -3.87    \\
\multicolumn{1}{c|}{}                        & \space    & \multicolumn{1}{c|}{\checkmark}    & 74.26     & \textcolor{ForestGreen}{+1.76}    \\ \hline
\multicolumn{1}{l|}{\multirow{4}{*}{ShuffleNet-V1}} & \space    & \multicolumn{1}{c|}{\space}    & 70.50     & -    \\
\multicolumn{1}{c|}{}                        & \checkmark    & \multicolumn{1}{c|}{\checkmark}    & 74.29     & \textcolor{ForestGreen}{+3.79}    \\
\multicolumn{1}{c|}{}                        & \checkmark    & \multicolumn{1}{c|}{\space}    & 70.52     & +0.02    \\
\multicolumn{1}{c|}{}                        & \space    & \multicolumn{1}{c|}{\checkmark}    & 74.91     & \textcolor{ForestGreen}{+4.41} \\ \hline
\multicolumn{5}{c}{\textit{WRN-40-2 as the teacher}}                                                          \\ \hline
\multicolumn{1}{c|}{\multirow{4}{*}{WRN-16-2}} & \space    & \multicolumn{1}{c|}{\space}    & 73.26     & -    \\
\multicolumn{1}{c|}{}                        & \checkmark    & \multicolumn{1}{c|}{\checkmark}    & 74.96     & \textcolor{ForestGreen}{+1.70}    \\
\multicolumn{1}{c|}{}                        & \checkmark    & \multicolumn{1}{c|}{\space}    & 70.96     & -2.30    \\
\multicolumn{1}{c|}{}                       & \space    & \multicolumn{1}{c|}{\checkmark}    & 74.76     & \textcolor{ForestGreen}{+1.50}    \\ \hline
\multicolumn{1}{c|}{\multirow{4}{*}{ShuffleNet-V1}} & \space    & \multicolumn{1}{c|}{\space}    & 70.50     & -    \\
\multicolumn{1}{c|}{}                        & \checkmark    & \multicolumn{1}{c|}{\checkmark}    & 74.92     & \textcolor{ForestGreen}{+4.42}    \\
\multicolumn{1}{c|}{}                        & \checkmark    & \multicolumn{1}{c|}{\space}    & 70.62     & +0.12    \\
\multicolumn{1}{c|}{}                        & \space    & \multicolumn{1}{c|}{\checkmark}    & 75.12     & \textcolor{ForestGreen}{+4.62}  \\ \hline
\end{tabular}
\end{small}
\vspace{-8pt} %tablespace
\caption{
	Accuracy(\%) on the CIFAR-100 validation set. $\Delta$ represents the performance improvement over the baseline.
	}
\label{tab:tckd_nckd}
\end{table}

Intuitively, TCKD concentrates on the knowledge related to the target class since the corresponding loss function considers only binary probabilities. Conversely, NCKD focuses on the knowledge among non-target classes. 
We notice that singly applying TCKD could be unhelpful~(\eg, $0.02\%$ and $0.12\%$ gain on ShuffleNet-V1) or even harmful~(\eg, $2.30\%$ drop on WRN-16-2 and $3.87\%$ drop on ResNet8$\times$4) for the student. However, the distillation performances of NCKD are comparable and even better than the classical KD~(\eg, $1.76\%$ \vs $1.13\%$ on ResNet8$\times$4). 
The ablation results suggest that the target-class-related knowledge could not be as important as knowledge among non-target classes. To dive into this phenomenon, we provide further analyses presented as follows.

% TCKD是在传递hardness相关的knowledge
\vspace{5pt}
\noindent \textbf{TCKD transfers the knowledge concerning the ``difficulty'' of training samples.}
According to Eqn.(\ref{form_dkd}), TCKD transfers ``dark knowledge'' via the binary classification task, which could be related to the sample ``difficulty''. For instance, a training sample with $p_t^{\mathcal{T}}=0.99$ could be ``easier'' for the student to learn compared with another one with $p_{t}^{\mathcal{T}}=0.75$. 
Since TCKD conveys the ``difficulty'' of training samples, we suppose the effectiveness would be revealed when the training data becomes challenging. However, the CIFAR-100 training set is easy to fit\footnote{Training accuracies on CIFAR-100 could be 100\% after convergence.}. Thus the knowledge of ``difficulty'' provided by the teacher is not informative. 
In this part, experiments from three perspectives are performed to validate: \textit{The more difficult the training data is, the more benefits TCKD could provide}\footnote{All experiments from these perspectives are performed with NCKD, since we suppose that TCKD should not be singly employed according to the results in Table~\ref{tab:tckd_nckd}. The probable reasons and analyzes are attached in the supplement.}.

\textit{(1) Applying Strong Augmentation} is a straightforward way to increase the difficulty of training data. We train a ResNet32$\times$4 model as the teacher with AutoAugment\cite{autoaug} on CIFAR-100, achieving 81.29\% top-1 validation accuracy. As for students, we train ResNet8$\times$4 and ShuffleNet-V1 models with/without TCKD. Results in Table~\ref{tab:auto_aug} reveal that TCKD obtains significant performance gains if strong augmentations are applied.

\begin{table}[h]
\setlength{\belowcaptionskip}{-10pt} %endtabspace
\center
\begin{small}
\begin{tabular}{c|c|cc}
student & \multicolumn{1}{c|}{TCKD} & \multicolumn{1}{c}{top-1} & $\Delta$ \\ \Xhline{3\arrayrulewidth}
\multirow{2}{*}{ResNet8$\times$4}       & \space                         & 73.82                     & -     \\
        & \checkmark                         & 75.33                     & \textcolor{ForestGreen}{+1.51} \\ \hline
\multirow{2}{*}{ShuffleNet-V1}        & \space                        & 77.13                     & -     \\
        & \checkmark                         & 77.98                     & \textcolor{ForestGreen}{+0.85}

\end{tabular}
\end{small}
\vspace{-8pt} %tablespace
\caption{Accuracy(\%) on the CIFAR-100 validation. We set ResNet32$\times$4 as the teacher and ResNet8$\times$4 as the student. Both teachers and students are trained with AutoAugment\cite{autoaug}.}
\label{tab:auto_aug}
\end{table}

% \vspace{-15pt} % table 2&3 space
\begin{table}[h]
\setlength{\belowcaptionskip}{-10pt} %endtabspace
\center
\begin{small}
\begin{tabular}{c|c|cc}
noisy ratio & \multicolumn{1}{c|}{TCKD} & \multicolumn{1}{c}{top-1} & $\Delta$ \\ \Xhline{3\arrayrulewidth}
\multirow{2}{*}{0.1}       & \space                         & 70.99                     & -     \\
        & \checkmark                         & 70.96                     & -0.03 \\ \hline
\multirow{2}{*}{0.2}        & \space                        & 67.55                     & -     \\
        & \checkmark                         & 68.03                     & \textcolor{ForestGreen}{+0.48} \\
        \hline
\multirow{2}{*}{0.3}        & \space                        & 64.62                     & -     \\
        & \checkmark                         & 65.26                     & \textcolor{ForestGreen}{+0.64}

\end{tabular}
\end{small}
\vspace{-8pt} %tablespace
\caption{Accuracy(\%) on the CIFAR-100 validation with different noisy ratios on the training set. We set ResNet32$\times$4 as the teacher and ResNet8$\times$4 as the student.}
\label{tab:nosiy_label}
\end{table}

\textit{(2) Noisy Labels} can also increase the difficulty of training data.
We train ResNet32$\times$4 models as teachers and ResNet8$\times$4 as students on CIFAR-100 with \{0.1, 0.2, 0.3\} symmetric noisy ratios, following \cite{nosiy_label2,coteach}. As reported in Table~\ref{tab:nosiy_label}, the results indicate that TCKD achieves more performance promotions on noisier training data.

\textit{(3) Challenging Datasets}~(\eg, ImageNet\cite{imagenet}) are also considered.
It shows that TCKD could bring $+0.32\%$ performance gain on ImageNet in Table~\ref{tab:imagenet_sec3}.

\begin{table}[h]
\setlength{\belowcaptionskip}{-10pt} %endtabspace
\center
\begin{small}
\begin{tabular}{c|cc}
\multicolumn{1}{c|}{TCKD} & \multicolumn{1}{c}{top-1} & $\Delta$ \\ \Xhline{3\arrayrulewidth}
\space                         & 70.71                     & -     \\
\checkmark                         & 71.03                     & \textcolor{ForestGreen}{+0.32} 

\end{tabular}
\end{small}
\vspace{-8pt} %tablespace
\caption{Accuracy(\%) on the ImageNet validation. We set ResNet-34 as the teacher and ResNet-18 as the student.}
\label{tab:imagenet_sec3}
\end{table}

Conclusively, we demonstrate the effectiveness of TCKD by experimenting with various strategies to increase the difficulty of training data~(\eg strong augmentation, noisy labels, difficult tasks). The results validate that the knowledge concerning the ``difficulty'' of training samples could be more beneficial when distilling knowledge on more challenging training data.

% NCKD被抑制了
\vspace{5pt}
\noindent \textbf{NCKD is the prominent reason why logit distillation works but is greatly suppressed.}
Interestingly, we notice in Table~\ref{tab:tckd_nckd} when only NCKD is applied, the performances are comparable or even better than the classical KD. 
It shows that the knowledge among non-target classes is of vital importance to logit distillation, which can be the prominent ``dark knowledge''. 
However, by reviewing Eqn.(\ref{form_dkd}), we notice that the NCKD loss is coupled with $(1-p_{t}^{\mathcal{T}})$, where $p_{t}^{\mathcal{T}}$ represents the teacher's confidence on the target class. Therefore, more confident predictions result in smaller NCKD weights. 
We suppose that \textit{the more confident the teacher is in the training sample, the more reliable and valuable knowledge it could provide.}
However, the loss weights are highly suppressed by such confident predictions. We suppose that this fact would limit the effectiveness of knowledge transfer, which is firstly investigated thanks to our reformulation of KD in Eqn.~(\ref{form_dkd}).

We design an ablation experiment to verify that well-predicted samples do transfer better knowledge than the others. Firstly we rank the training samples according to $p_{t}^{\mathcal{T}}$, and evenly split them into two sub-sets. For clarity, one sub-set includes samples with top-50\% $p_{t}^{\mathcal{T}}$ while remaining samples are in the other sub-set. Then we train student networks with NCKD on each subset to compare the performance gain~(while the cross-entropy loss is still on the whole set). Table~\ref{tab:top_bottom_nckd} shows that utilizing NCKD on the top-50\% samples achieves better performance, suggesting that the knowledge of well-predicted samples is richer than others. However, the loss weight of well-predicted samples are suppressed by the high confidence of the teacher.

% \begin{table}[h]
% \setlength{\belowcaptionskip}{-10pt} %endtabspace
% \center
% \begin{small}
% \begin{tabular}{cc|ccc}
% 0-50\% & \multicolumn{1}{c|}{50-100\%} & \multicolumn{1}{c}{$1 - p_{t}^{\mathcal{T}}$} & top-1\\ \Xhline{3\arrayrulewidth}
% \checkmark        & \checkmark      & 0.76             & 74.26     \\
% \checkmark        & \space          & 0.68             & 74.23     \\
% \space            & \checkmark      & 0.83             & 73.96     \\
% \end{tabular}
% \end{small}
% \vspace{-8pt} %tablespace
% \caption{Accuracy(\%) on the CIFAR-100 validation set. We set ResNet32$\times$4 as the teacher and ResNet8$\times$4 as the student.}
% \label{tab:top_bottom_nckd}
% \end{table}

\begin{table}[h]
\setlength{\belowcaptionskip}{-10pt} %endtabspace
\center
\begin{small}
\begin{tabular}{cc|ccc}
0-50\% & \multicolumn{1}{c|}{50-100\%} & top-1\\ \Xhline{3\arrayrulewidth}
\checkmark        & \checkmark                   & 74.26     \\
\checkmark        & \space                       & 74.23     \\
\space            & \checkmark                  & 73.96     \\
\end{tabular}
\end{small}
\vspace{-8pt} %tablespace
\caption{Accuracy(\%) on the CIFAR-100 validation set. We set ResNet32$\times$4 as the teacher and ResNet8$\times$4 as the student.}
\label{tab:top_bottom_nckd}
\end{table}

% 提出dkd，解决我们发现的两个问题
\subsection{Decoupled Knowledge Distillation}
So far, we have reformulated the classical KD loss into a weighted sum of two independent parts, and further validated the effectiveness of TCKD and revealed the suppression of NCKD. 
Specifically, TCKD transfers knowledge concerning the ``difficulty'' of training samples. More significant improvements could be obtained by TCKD on more challenging training data.
NCKD transfers knowledge among non-target classes, which would be suppressed in the condition that the weight $(1-p_{t}^{\mathcal{T}})$ is relatively small.

Instinctively, both TCKD and NCKD are indispensable and crucial. However, in the classical KD formulation, TCKD and NCKD are coupled from the following aspects:
\begin{itemize}
\setlength{\itemsep}{0pt}
\setlength{\parsep}{0pt}
\setlength{\parskip}{0pt}
	\item For one thing, NCKD is coupled with $(1-p_{t}^{\mathcal{T}})$, which could suppress NCKD on the well-predicted samples. Since results in Table~\ref{tab:top_bottom_nckd} show that well-predicted samples could bring more performance gain, the coupled form could limit the effectiveness of NCKD. 
	\item For another, weights of NCKD and TCKD are coupled under the classical KD framework. It's not allowed to change each term's weight to balance the importance. We suppose that TCKD and  NCKD should be separately considered since their contributions are from different aspects. % which we suppose should be independently considered. 
\end{itemize}

\begin{algorithm}[t]
    \caption{Pseudo code of DKD in a PyTorch-like style.}
    \label{algo:ctc}
    \footnotesize
    \begin{alltt}
    \color{ForestGreen}
# l_stu: student output logits
# l_tea: teacher output logits
# T: temperature for KD & DKD
# t: labels, (N, C), bool type
# alpha, beta: hyper-parameters for DKD
\end{alltt}
\vspace{-15pt}
\begin{alltt}
p_stu = F.softmax(l_stu / T)
p_tea = F.softmax(l_tea / T)
\color{ForestGreen}# pt & pnt: (N, 1), Eqn.(2) \color{Black}
pt_stu, pnt_stu = p_stu[t], p_stu[1-t].sum(1)
pt_tea, pnt_tea = p_tea[t], p_tea[1-t].sum(1)
\color{ForestGreen}# pnct: (N, C-1), Eqn.(3) \color{Black}
pnct_stu = F.softmax(l_stu[1-t]/T)
pnct_tea = F.softmax(l_tea[1-t]/T)
    
\color{ForestGreen}# TCKD \color{Black}
tckd = kl_div(log(pt_stu), pt_tea) + 
       kl_div(log(pnt_stu), pnt_tea)
\color{ForestGreen}# NCKD \color{Black}
nckd = F.kl_div(log(pnct_stu), pnct_tea)
    
\color{ForestGreen}# ori KD \color{Black}
\color{ForestGreen}# kd_loss = (tckd + pnt_tea*nckd) * T**2 \color{Black}
\color{ForestGreen}# DKD \color{Black}
dkd_loss = (alpha*tckd + beta*nckd) * T**2  
\end{alltt}
\end{algorithm}

Benefiting from our reformulation of KD, we propose a novel logit distillation method named Decoupled Knowledge Distillation(DKD) to address the above issues.
Our proposed DKD independently considers TCKD and NCKD in a decoupled formulation, as shown in Figure~\ref{fig1b}. Specifically, we introduce two hyper-parameters $\alpha$ and $\beta$, as the weights of TCKD and NCKD, respectively. The loss function of DKD can be written as follows:
\begin{equation}
	\text{DKD} = \alpha \text{TCKD} + \beta \text{NCKD}.
\end{equation}
In DKD, $(1-p_{t}^{\mathcal{T}})$, which would suppress NCKD's effectiveness, is replaced by $\beta$. What's more, it's allowed to adjust $\alpha$ and $\beta$ to balance the importance of TCKD and NCKD. Through decoupling NCKD and TCKD, DKD provides an efficient and flexible manner for logit distillation.
Algorithm~\ref{algo:ctc} provides the pseudo-code of DKD in a PyTorch-like\cite{torch} style. 

\begin{table*}[h]
\setlength{\belowcaptionskip}{-10pt} %endtabspace
\center
\begin{small}
\begin{tabular}{cc|cccccc}
\multirow{4}{*}{\begin{tabular}[c]{@{}c@{}}distillation \\ manner\end{tabular}} & \multirow{2}{*}{teacher}  & ResNet56 & ResNet110 & ResNet32$\times$4 & WRN-40-2 & WRN-40-2 & VGG13 \\
&      & 72.34      & 74.31       & 79.42        & 75.61      & 75.61      & 74.64   \\
& \multirow{2}{*}{student}  & ResNet20 & ResNet32 & ResNet8$\times$4  & WRN-16-2 & WRN-40-1 & VGG8  \\
& \space     & 69.06      & 71.14       & 72.50        & 73.26      & 71.98      & 70.36   \\ \Xhline{3\arrayrulewidth} 
\multirow{5}{*}{features}
& FitNet\cite{fitnets}   & 69.21      & 71.06       & 73.50        & 73.58      & 72.24      & 71.02   \\
& RKD\cite{rkd}      & 69.61      & 71.82       & 71.90        & 73.35      & 72.22      & 71.48   \\
& CRD\cite{crd}      & 71.16      & 73.48       & 75.51        & 75.48      & 74.14      & 73.94   \\
& OFD\cite{ofd}      & 70.98      & 73.23       & 74.95        & 75.24      & 74.33      & 73.95   \\
& ReviewKD\cite{revkd} & 71.89      & 73.89       & 75.63        & 76.12      & \textbf{75.09}      & \textbf{74.84}   \\ \hline 
\multirow{3}{*}{logits}                                                         
& KD\cite{kd}       & 70.66      & 73.08       & 73.33        & 74.92      & 73.54      & 72.98   \\
& \textbf{DKD}      & \textbf{71.97}      & \textbf{74.11}       & \textbf{76.32}        & \textbf{76.24}      & 74.81      & 74.68 \\
& $\Delta$      & \textcolor{ForestGreen}{+1.31}      & \textcolor{ForestGreen}{+1.03}       & \textcolor{ForestGreen}{+2.99}        & \textcolor{ForestGreen}{+1.32}      & \textcolor{ForestGreen}{+1.27}      & \textcolor{ForestGreen}{+1.70} \\
\end{tabular}
\end{small}
\vspace{-8pt} %tablespace
\caption{\textbf{Results on the CIFAR-100 validation.} Teachers and students are in the \textbf{same} architectures. And $\Delta$ represents the performance improvement over the classical KD. All results are the average over 5 trials.}
\label{tab:xifa}
\end{table*}

\section{Experiments}
\label{sec:exp}
We mainly experiment on two representative tasks, \ie, image classification and object detection, including:

\textit{\textbf{CIFAR-100}\cite{cifar}} is a well-known image classification dataset, containing $32\times 32$ images of 100 categories. Training and validate sets are composed of 50k and 10k images.

\textit{\textbf{ImageNet}\cite{imagenet}} is a large-scale classification dataset that consists of 1000 classes. The training set contains 1.28 million images and the validation set contains 50k images.

\textit{\textbf{MS-COCO}\cite{coco}} is an 80-category general object detection dataset. The \texttt{train2017} split contains 118k images, and the \texttt{val2017} split contains 5k images.

All implementation details are attached in supplement due to the page limit.

\subsection{Main Results}

Firstly, we demonstrate the improvements contributed by decoupling (1) NCKD and $p_t^{\mathcal{T}}$ and (2) NCKD and TCKD, respectively. Then, we benchmark our method on image classification and object detection tasks.

\vspace{5pt}
\noindent \textbf{Ablation: $\alpha$ and $\beta$.}
The two tables below report the student accuracy~(\%) with different $\alpha$ and $\beta$. ResNet32$\times$4 and ResNet8$\times$4 are set as the teacher and the student, respectively. Firstly, we prove that decoupling $(1-p_{t}^{\mathcal{T}})$ and NCKD can bring reasonable performance gain~(73.63\% \vs 74.79\%) in the first table. Then, we demonstrate that decoupling weights of NCKD and TCKD could contribute to further improvements~(74.79\% \vs 76.32\%). Moreover, the second table indicates that TCKD is indispensable, and the improvements from TCKD are stable with different $\alpha$ around 1.0\footnote{We fix $\alpha$ as 1.0 for simplification in the first table, and $\beta$ as 8.0 in the second table since it achieves the best performance in the first one.}.

\begin{table}[h]
\setlength{\belowcaptionskip}{-10pt} %endtabspace
\center
\begin{small}
\begin{tabular}{c|c|ccccc}
$\beta$  & $1-p_{t}^{\mathcal{T}}$ & 1.0 & 2.0 & 4.0 & 8.0 & 10.0 \\ \Xhline{3\arrayrulewidth}
top-1 & 73.63 & 74.79 & 75.44 & 75.94 & \textbf{76.32} & 76.18
\end{tabular}

\begin{tabular}{c|cccccc}
$\alpha$  & 0.0 & 0.2 & 0.5 & 1.0 & 2.0 & 4.0 \\ \Xhline{3\arrayrulewidth}
top-1 & 75.30 & 75.64 & 76.12 & \textbf{76.32} & 76.11 & 75.42
\end{tabular}
\end{small}
\end{table}

\begin{table*}[th]
\setlength{\belowcaptionskip}{-10pt} %endtabspace
\center
\begin{small}
\begin{tabular}{cc|ccccc}
\multirow{4}{*}{\begin{tabular}[c]{@{}c@{}}distillation \\ manner\end{tabular}} & \multirow{2}{*}{teacher}  & ResNet32$\times$4 & WRN-40-2 & VGG13 & ResNet50 & ResNet32$\times$4 \\
& \space     & 79.42      & 75.61       & 74.64        & 79.34      & 79.42         \\
& \multirow{2}{*}{student}  & ShuffleNet-V1 & ShuffleNet-V1 & MobileNet-V2  & MobileNet-V2 & ShuffleNet-V2  \\
& \space     & 70.50      & 70.50       & 64.60        & 64.60      & 71.82         \\ \Xhline{3\arrayrulewidth} 
\multirow{5}{*}{features}
& FitNet\cite{fitnets} & 73.59      & 73.73       & 64.14        & 63.16      & 73.54         \\
& RKD\cite{rkd}      & 72.28      & 72.21       & 64.52        & 64.43      & 73.21         \\
& CRD\cite{crd}      & 75.11      & 76.05       & 69.73        & 69.11      & 75.65         \\
& OFD\cite{ofd}      & 75.98      & 75.85       & 69.48        & 69.04      & 76.82         \\
& ReviewKD\cite{revkd} & \textbf{77.45}      & \textbf{77.14}       & \textbf{70.37}        & 69.89      & \textbf{77.78}         \\ \hline 
\multirow{3}{*}{logits}                                                         
& KD\cite{kd}       & 74.07      & 74.83       & 67.37        & 67.35      & 74.45         \\
& \textbf{DKD}      & 76.45      & 76.70       & 69.71        & \textbf{70.35}      & 77.07       \\
& $\Delta$      & \textcolor{ForestGreen}{+2.38}      & \textcolor{ForestGreen}{+1.87}       & \textcolor{ForestGreen}{+2.34}        & \textcolor{ForestGreen}{+3.00}      & \textcolor{ForestGreen}{+2.62} \\
\end{tabular}
\end{small}
\vspace{-8pt} %tablespace
\caption{\textbf{Results on the CIFAR-100 validation.} Teachers and students are in \textbf{different} architectures. And $\Delta$ represents the performance improvement over the classical KD. All results are the average over 5 trials.}
\label{tab:xifa2}
\end{table*}

\begin{table*}[h]
\center
\begin{small}
\begin{tabular}{ccc|cccc|ccc}
\multicolumn{3}{c|}{distillation manner} & \multicolumn{4}{c|}{features}    & \multicolumn{3}{c}{logits} \\ \Xhline{3\arrayrulewidth}
           & teacher      & student     & AT\cite{at}    & OFD\cite{ofd}   & CRD\cite{crd}   & ReviewKD\cite{revkd} & KD\cite{kd} & KD*           & \textbf{DKD}          \\ \hline
top-1      & 73.31        & 69.75       & 70.69 & 70.81 & 71.17 & 71.61    & 70.66 & 71.03 & \textbf{71.70}        \\
top-5      & 91.42        & 89.07       & 90.01 & 89.98 & 90.13 & \textbf{90.51}    & 89.88 & 90.05 & 90.41       \\ 
\end{tabular}
\vspace{-8pt} %tablespace
\caption{\textbf{Top-1 and top-5 accuracy~(\%) on the ImageNet validation.} We set \textbf{ResNet-34} as the teacher and \textbf{ResNet-18} as the student. KD* represents the result of our implementation. All results are the average over 3 trials.}
\label{tab:yingmeijunaite}

\setlength{\belowcaptionskip}{-10pt} %endtabspace
\begin{tabular}{ccc|cccc|ccc}
\multicolumn{3}{c|}{distillation manner} & \multicolumn{4}{c|}{features}    & \multicolumn{3}{c}{logits} \\ \Xhline{3\arrayrulewidth}
           & teacher      & student     & AT\cite{at}    & OFD\cite{ofd}   & CRD\cite{crd}   & ReviewKD\cite{revkd} & KD\cite{kd}  & KD*         & \textbf{DKD}          \\ \hline
   top-1      & 76.16        & 68.87       & 69.56 & 71.25 & 71.37 & \textbf{72.56}    & 68.58  & 70.50     & 72.05         \\
top-5      & 92.86        & 88.76       & 89.33 & 90.34 & 90.41 & 91.00    & 88.98 &  89.80     &   \textbf{91.05}  
\end{tabular}
\end{small}
\vspace{-8pt} %tablespace
\caption{\textbf{Top-1 and top-5 accuracy~(\%) on the ImageNet validation.} We set \textbf{ResNet-50} as the teacher and \textbf{MobileNet-V1} as the student. KD* represents the result of our implementation. All results are the average over 3 trials.}
\label{tab:yingmeijunaite2}
\end{table*}

\vspace{5pt}
\noindent \textbf{CIFAR-100 image classification.}
We discuss experimental results on CIFAR-100 to examine our DKD. The validation accuracy is reported in Table~\ref{tab:xifa} and Table~\ref{tab:xifa2}. Table~\ref{tab:xifa} contains the results where teachers and students are of the same network architectures. Table~\ref{tab:xifa2} shows the results where teachers and students are from different series. 

Notably, DKD achieves consistent improvements on all teacher-student pairs, compared with the baseline and the classical KD. Our method achieves $1\sim 2\%$ and $2\sim 3\%$ improvements on teacher-student pairs of the same and different series, respectively. It strongly supports the effectiveness of DKD.
Furthermore, DKD achieves comparable or even better performances than feature-based distillation methods, significantly improving the trade-off between distillation performance and training efficiency, which will be further discussed in Sec~\ref{sec:ext}. 

\vspace{5pt}
\noindent \textbf{ImageNet image classification.}
Top-1 and top-5 accuracies of image classification on ImageNet are reported in Table~\ref{tab:yingmeijunaite} and Table~\ref{tab:yingmeijunaite2}.
Our DKD achieves a significant improvement. It's worth mentioning that the performance of DKD is better than the most state-of-the-art results of feature distillation methods.
% 1. 同样的，超过了KD的performance and 达到了和feature蒸馏差不多的效果；

\vspace{5pt}
\noindent \textbf{MS-COCO object detection.}
% 1. Detection任务必须依赖feature蒸馏，因为依赖location信息，只蒸馏logits是无法有效提供knowledge的，所以我们提供了logits+feature蒸馏结合的results
% 2. 单独使用DKD，也是consistently超过了KD的结果的。在review KD的高baseline基础上还可以继续涨，达到了SOTA
% 3. 详细讲一讲超过了几个点
As discussed in previous works, the performance of the object detection task greatly depends on the quality of deep features to locate interested objects. This rule also stands in transferring knowledge between detectors~\cite{detmimick1,detmimick2}, \ie, feature mimicking is of vital importance since logits are not capable of providing knowledge for object localization. As shown in Table~\ref{tab:coco}, singly applying DKD can hardly achieve outstanding performances, but expectedly surpasses the classical KD. Thus, we introduce the feature-based distillation method ReviewKD\cite{revkd} to obtain satisfactory results. It can be observed that our DKD can further boost AP metrics, even the distillation performance of ReviewKD is relatively high. Conclusively, new state-of-the-art results are obtained by combining our DKD with feature-based distillation methods on the object detection task.

\begin{table*}[h]
\setlength{\belowcaptionskip}{-10pt} %endtabspace
\center
\begin{small}
\begin{tabular}{c|ccc|ccc|ccc}
\multicolumn{1}{c|}{} & \multicolumn{3}{c|}{R-101 \& R-18} & \multicolumn{3}{c|}{R-101 \& R-50} & \multicolumn{3}{c}{R-50 \& MV2} \\
\space                & AP         & AP$_{50}$      & AP$_{75}$      & AP         & AP$_{50}$      & AP$_{75}$      & AP        & AP$_{50}$      & AP$_{75}$      \\ \Xhline{3\arrayrulewidth}
teacher               & 42.04      & 62.48     & 45.88     & 42.04      & 62.48     & 45.88     & 40.22     & 61.02     & 43.81     \\
student               & 33.26      & 53.61     & 35.26     & 37.93      & 58.84     & 41.05     & 29.47     & 48.87     & 30.90     \\ \hline
KD\cite{kd}                    & 33.97      & 54.66     & 36.62     & 38.35      & 59.41     & 41.71     & 30.13     & 50.28     & 31.35     \\
FitNet\cite{fitnets}                & 34.13      & 54.16     & 36.71     & 38.76      & 59.62     & 41.80     & 30.20     & 49.80     & 31.69     \\
FGFI\cite{fgfi}                  & 35.44      & 55.51     & 38.17     & 39.44      & 60.27     & 43.04     & 31.16     & 50.68     & 32.92     \\
ReviewKD\cite{revkd}              & 36.75      & 56.72     & 34.00     & 40.36      & 60.97     & 44.08     & 33.71     & 53.15     & 36.13     \\
\hline
\textbf{DKD}                   & 35.05      & 56.60     & 37.54     & 39.25      & 60.90     & 42.73     & 32.34     & 53.77     & 34.01     \\
\textbf{DKD+ReviewKD}          & \textbf{37.01}      & \textbf{57.53}     & \textbf{39.85}     & \textbf{40.65}      & \textbf{61.51}     & \textbf{44.44} & \textbf{34.35}      & \textbf{54.89}     & \textbf{36.61}    
\end{tabular}
\end{small}
\vspace{-8pt} %tablespace
\caption{\textbf{Results on MS-COCO based on Faster-RCNN\cite{faster_rcnn}-FPN\cite{fpn}}: AP evaluated on \texttt{val2017}. Teacher-student pairs are ResNet-101~(R-101) \& ResNet-18~(R-18), ResNet-101 \& ResNet-50~(R-50) and ResNet-50 \& MobileNet-V2~(MV2) respectively. All results are the average over 3 trials. More details are attached in supplement.}
\label{tab:coco}
\end{table*}

\subsection{Extensions}
\label{sec:ext}
For a better understanding of DKD, we provide extensions from four perspectives. First of all, we comprehensively compare the training efficiency of DKD with representative state-of-the-art methods. 
Then, we provide a new perspective to explain why \textit{bigger models are not always better teachers} and alleviate this problem by utilizing DKD. Moreover, following~\cite{crd}, we examine the transferability of deep features learned by DKD. And we also present some visualizations to validate the superiority of DKD.

\begin{figure}[h]
\setlength{\abovecaptionskip}{10pt} %figurespace
\setlength{\belowcaptionskip}{-10pt} %endfigspace
	\centering
    \includegraphics[width=0.9\linewidth]{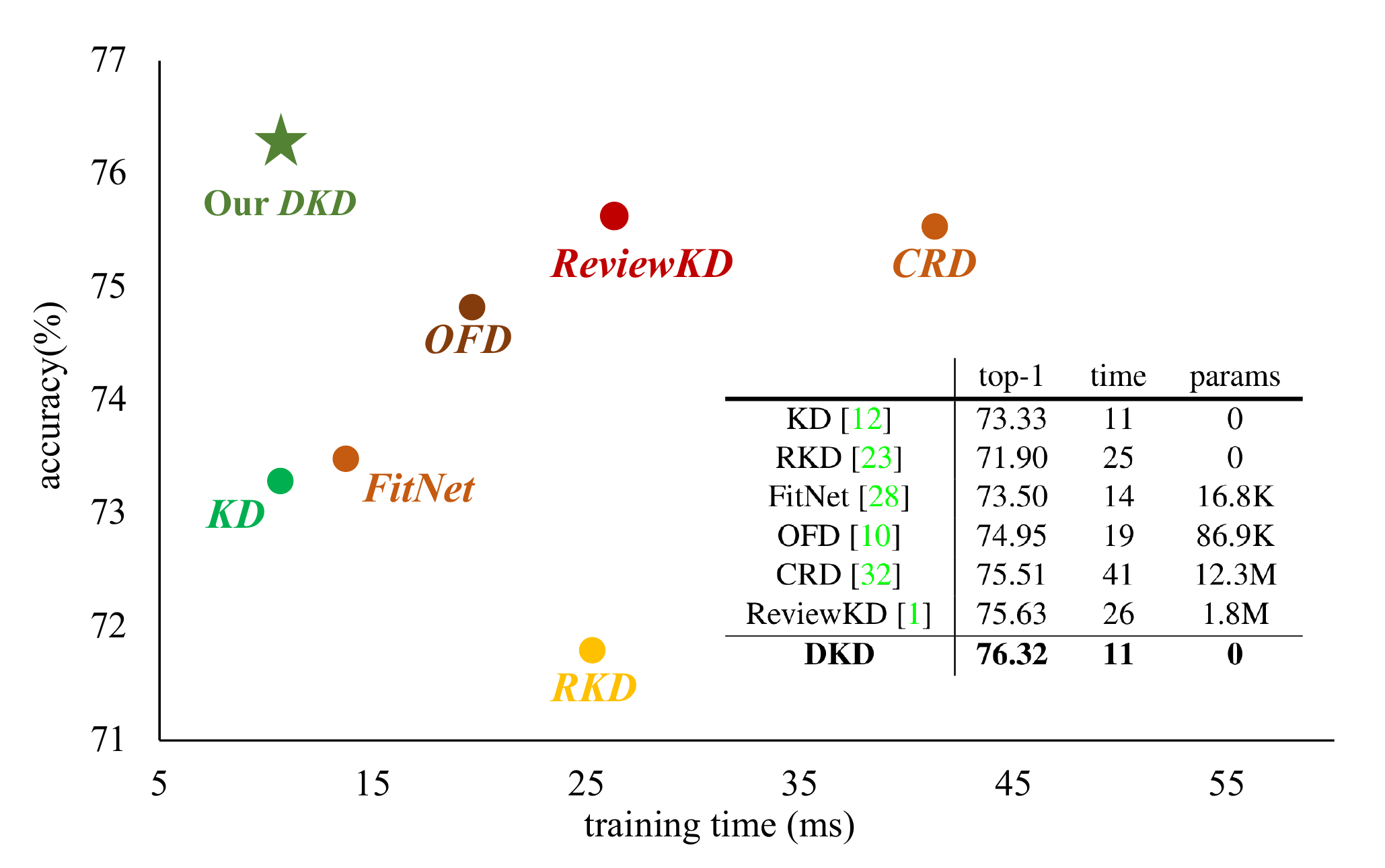}
% 	\label{speed}
	\caption{Training time~(per batch) \vs accuracy on CIFAR-100. We set ResNet32$\times$4 as the teacher and ResNet8$\times$4 as the student. The table shows the number of extra parameters for each method.}
	\label{fig:speed}
\end{figure}

\vspace{5pt}
\noindent \textbf{Training efficiency.}
% 4. training时间 or FLOPs的统计，和performance，画一个二维坐标
We assess the training costs of state-of-the-art distillation methods, proving the high training efficiency of DKD. 
As shown in Figure~\ref{fig:speed}, our DKD achieves the best trade-off between model performances and training costs~(\eg, training time and extra parameters). Since DKD is reformulated from the classical KD, it needs almost the same computational complexity as KD, and of course no extra parameters. However, feature-based distillation methods require extra training time for distillation intermediate layer features, as well as the GPU memory costs. 

\vspace{5pt}
\noindent \textbf{Improving performances of big teachers.}
% 1. 我们猜测的原因是，越大的teacher，NCKD被抑制更严重，实际上，它是可以transfer更有价值的knowledge的，实验也证明了至少不会变差；
% 2. 其他paper也提到了，比如使用ESKD和teaching assistant，实际上也是在缓解NCKD的抑制，ESKD选择了靠前epoch的teacher，此时模型并没有over-convergence，所以Pt的value会比较小，对NCKD的抑制会比较弱；同时，teaching assistant选择了更小的模型作为bridge，自然地，小模型产生的pt会更小，从而缓解NCKD的抑制；
% 3. 我们找到了一个更加general的解释，来解释为什么大模型不总是好teacher；
We provide a new potential explanation on the \textit{bigger models are not always better teachers} issue. Specifically, bigger teachers are expected but cannot transfer more beneficial knowledge, even achieving worse performances than smaller ones. 

Previous works\cite{eskd,survey} explained this phenomenon with the large capacity gap between big teachers and small students. However, we suppose that the main reason is the suppression of NCKD, \ie, the $(1-p_{t}^{\mathcal{T}})$ would become smaller with the teacher getting bigger. 
What's more, related works on this problem also could be explained from this perspective, \eg, ESKD\cite{eskd} employs early-stopped teacher models to alleviate this problem, and these teachers would be under-convergence and yield smaller $p_{t}^{\mathcal{T}}$. 

To validate our conjecture, we perform our DKD on a series of teacher models. Experimental results in Table~\ref{tab:big_tea} and Table~\ref{tab:big_tea2} consistently indicate that our DKD alleviates the \textit{bigger models are not always better teachers} problem.

\begin{table}[th]
\setlength{\belowcaptionskip}{-10pt} %endtabspace
\center
\begin{small}
\begin{tabular}{c|cccc}
\multirow{2}{*}{teacher} & W-28-2 & W-40-2 & W-16-4 & W-28-4 \\
                         & 75.45        & 75.61        & 77.51        & 78.60        \\ \Xhline{3\arrayrulewidth}
KD                       & 75.37        & 74.92        & 75.79        & 75.04        \\
DKD                      & 75.92        & 76.24        & 76.00        & 76.45       
\end{tabular}
\end{small}
\vspace{-8pt} %tablespace
\caption{Results on CIFAR-100. We set WRN-16-2 as the student and WRN series networks as teachers.}
\label{tab:big_tea}
\end{table}

\begin{table}[th]
\setlength{\belowcaptionskip}{-10pt} %endtabspace
\center
\begin{small}
\begin{tabular}{c|ccc}
\multirow{2}{*}{teacher} & VGG13 & WRN-16-4 & ResNet50 \\
                         & 74.64        & 77.51        & 79.34        \\ \Xhline{3\arrayrulewidth}
KD                       & 74.93        & 75.79        & 75.36                \\
DKD                      & 75.45        & 76.00        & 76.60              
\end{tabular}
\end{small}
\vspace{-8pt} %tablespace
\caption{Results on CIFAR-100. We set WRN-16-2 as the student and networks from different series as teachers.}
\label{tab:big_tea2}
\end{table}

\vspace{5pt}
\noindent \textbf{Feature transferability.}
% 1. 很多任务证明了KD是可以学到更加generalizable的knowledge的，我们也提供相应的实验来证明这一点
% 2. 交代setting，fix backbone在下游数据集上做linear probing；
% 3. 我们的performance在所有的方法中，达到了最优的transfer性能，进一步说明，强化non-target information transfer的学习能够使knowledge transfer更加高效
We perform experiments to evaluate the transferability of deep features to verify that our DKD transfers more generalizable knowledge. Following \cite{crd}, we use the WRN-16-2 distilled from WRN-40-2 as the feature extractor. Then linear probing tasks are performed on downstream datasets, \ie STL-10\cite{stl10} and Tiny-ImageNet\footnote{\href{https://www.kaggle.com/c/tiny-imagenet}{https://www.kaggle.com/c/tiny-imagenet}}. Results are reported in Table~\ref{tab:feat_trans}, proving the outstanding transferability of features learned with our DKD. Implementation details are in the supplement.
\begin{table}[h]
\vspace{-10pt}
\setlength{\belowcaptionskip}{-10pt} %endtabspace
\center
\begin{small}
\setlength{\tabcolsep}{1.2mm}{
\begin{tabular}{c|cccccc}
             & baseline & KD & FitNet & CRD  & ReviewKD & \textbf{DKD}  \\ \Xhline{3\arrayrulewidth}
STL-10       & 69.7    & 70.9 & 70.3   & 71.6 & 72.4 & \textbf{72.9} \\
TI & 33.7    & 33.9 & 33.5  & 35.6 & 36.6 & \textbf{37.1}
\end{tabular}}
\end{small}
\vspace{-8pt} %tablespace
\caption{\textbf{Comparison with previous methods on transferring features} from CIFAR-100 to STL-10 and Tiny-ImageNet~(TI).}
\label{tab:feat_trans}
\end{table}

\begin{figure}[h]
	\centering
    \includegraphics[width=0.43\linewidth]{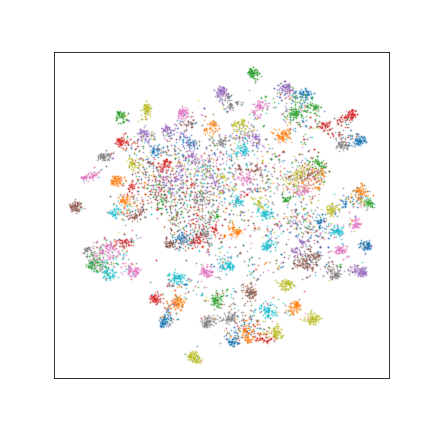}
    \includegraphics[width=0.43\linewidth]{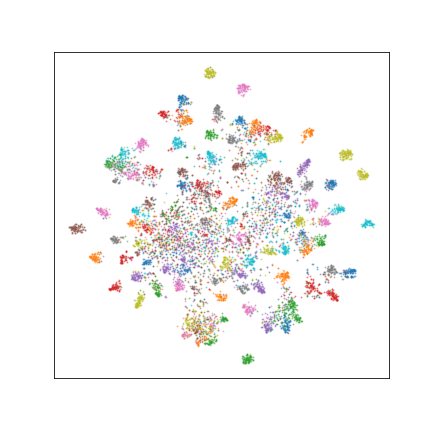}
    \vspace{-10pt}
	\caption{t-SNE of features learned by KD~(left) and DKD~(right).}
	\label{fig:tsne}
	\vspace{-15pt}
\end{figure}

\begin{figure}[h]
	\centering
    \includegraphics[width=0.43\linewidth]{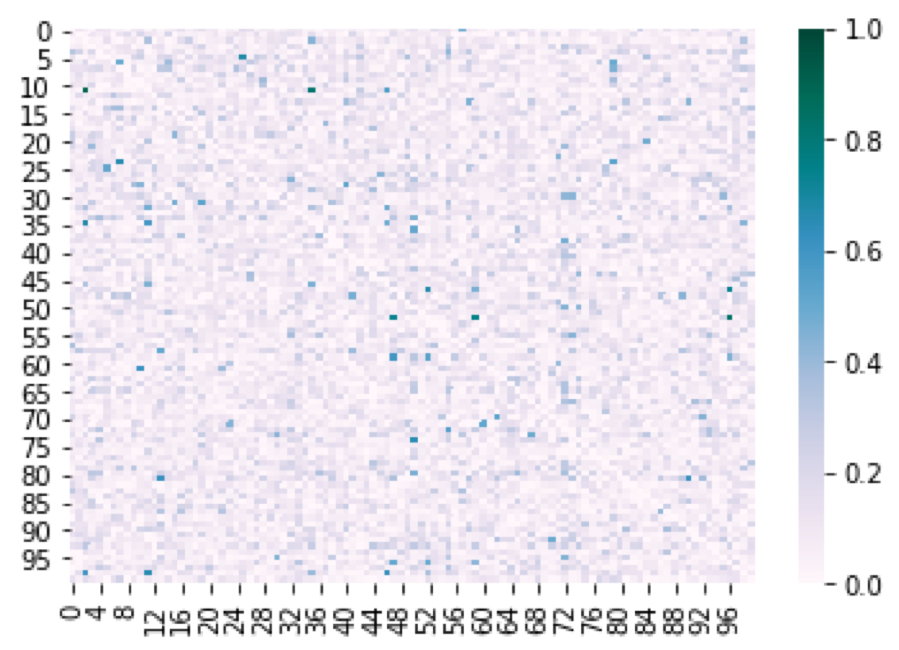}
    \includegraphics[width=0.43\linewidth]{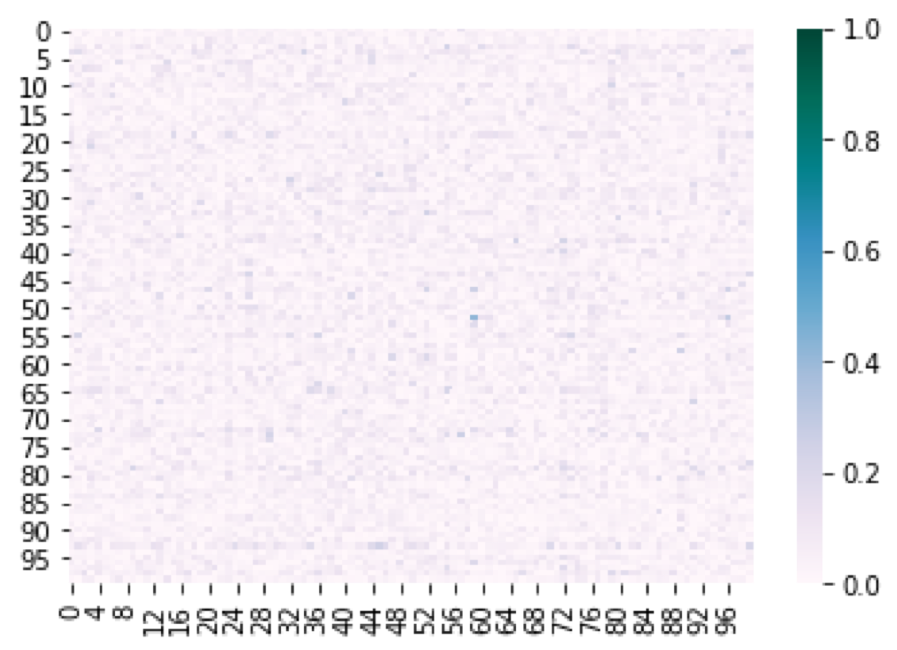}
    \vspace{-10pt}
	\caption{Difference of correlation matrices of student and teacher logits. Obviously, DKD~(right) leads to a smaller difference~(more similar prediction) than KD~(left).}
	\label{fig:diff}
	\vspace{-15pt}
\end{figure}

\vspace{5pt}
\noindent \textbf{Visualizations.}
We present visualizations from two perspectives~(with setting teacher as ResNet32x4 and student as ResNet8x4 on CIFAR-100).
(1) The t-SNE~(Fig.~\ref{fig:tsne}) results show that representations of DKD are more separable than KD, proving that DKD benefits the discriminability of deep features. (2) We also visualize the difference of correlation matrices of student and teacher logits~(Fig.~\ref{fig:diff}). Compared with KD, DKD helps the student to output more similar logits with the teacher, \ie, achieving better distillation performances.

\section{Discussion and Conclusion}
This paper provides a novel viewpoint to interpret logit distillation by reformulating the classical KD loss into two parts, \ie, target class knowledge distillation~(TCKD) and non-target class knowledge distillation~(NCKD). 
The effects of both parts are respectively investigated and proved.
More importantly, we reveal that the coupled formulation of KD limits the effectiveness and flexibility of knowledge transfer. 
To overcome these issues, we propose Decoupled Knowledge Distillation~(DKD), which achieves significant improvements on CIFAR-100, ImageNet and MS-COCO datasets for image classification and object detection tasks. Besides, the superiority of DKD in training efficiency and feature transferability is also demonstrated. We hope this paper will contribute to future logit distillation research.

\vspace{5pt}
\noindent \textbf{limitations and future works.} Noticeable limitations are discussed as follows. DKD could not outperform state-of-the-art feature-based methods~(\eg, ReviewKD~\cite{revkd}) on object detection tasks because logits-based methods cannot transfer knowledge about localization. Besides, we have provided an intuitive guidance on how to adjust $\beta$ in our supplement. However, the strict correlation between the distillation performance and $\beta$ is not fully investigated, which will be our future research direction.

%%%%%%%%% REFERENCES
{\small
\bibliographystyle{ieee_fullname}
\bibliography{egbib}
}

%%%%%%%%%%% Supp
%%%%%%%%%%%
%%%%%%%%%%%

\clearpage
%\newpage
\appendix       %%% starting appendix
\section*{A. Appendix}
\setcounter{table}{0}
\renewcommand{\thetable}{A.\arabic{table}}
% \section*{Appendix}

\subsection*{A.1. Details about the reformulation in Sec~\ref{sec:reform_kd}}

Details of the mathematical derivation in Sec~\ref{sec:reform_kd} of the manuscript are as follows~(notations are the same in Sec~\ref{sec:reform_kd} of the manuscript):

\vspace{-1pt}
\begin{equation}
    \begin{split}
    \text{KD} &= \text{KL}(\mathbf{p}^{\mathcal{T}}|| \mathbf{p}^{\mathcal{S}})\\
     &= \sum_{i=1}^{C} p^{\mathcal{T}}_{i}\log(\frac{p^{\mathcal{T}}_{i}}{p^{\mathcal{S}}_{i}})\\
    &=p^{\mathcal{T}}_{t}\log(\frac{p^{\mathcal{T}}_{t}}{p^{\mathcal{S}}_{t}}) + \sum_{i=1,i\neq t}^{C} p^{\mathcal{T}}_{i}\log(\frac{p^{\mathcal{T}}_{i}}{p^{\mathcal{S}}_{i}}).
    \end{split}
    \label{sup_reform_detail1}
\end{equation}
According to Eqn.(\ref{defpi}) and Eqn.(\ref{defP}) of the manuscript, we have $\hat{p}_{i}=p_{i}/p_{\backslash t}$. Thus, we can rewrite Eqn.(\ref{sup_reform_detail1}) to:
\vspace{-1pt}
\begin{small}
\begin{equation}
	\begin{split}
	\text{KD} &= p^{\mathcal{T}}_{t} \log(\frac{p^{\mathcal{T}}_{t}}{p^{\mathcal{S}}_{t}})+\sum_{i=1, i\neq t}^{C} p^{\mathcal{T}}_{\backslash t} \hat p^{\mathcal{T}}_{i}\log(\frac{p^{\mathcal{T}}_{\backslash t}\hat p^{\mathcal{T}}_{i}}{p^{\mathcal{S}}_{\backslash t}\hat p^{\mathcal{S}}_{i}})\\
	&= p^{\mathcal{T}}_{t}\log(\frac{p^{\mathcal{T}}_{t}}{p^{\mathcal{S}}_{t}}) + \sum_{i=1, i\neq t}^{C} p^{\mathcal{T}}_{\backslash t} \hat p^{\mathcal{T}}_{i}(\log(\frac{\hat p^{\mathcal{T}}_{i}}{\hat p^{\mathcal{S}}_{i}}) +\log(\frac{p^{\mathcal{T}}_{\backslash t}}{p^{\mathcal{S}}_{\backslash t}})) \\
	&= p^{\mathcal{T}}_{t} \log(\frac{p^{\mathcal{T}}_{t}}{p^{\mathcal{S}}_{t}}) + \sum_{i=1,i\neq t}^{C} p^{\mathcal{T}}_{\backslash t} \hat p^{\mathcal{T}}_{i} \log(\frac{\hat p^{\mathcal{T}}_{i}}{\hat p^{\mathcal{S}}_{i}}) \\
	&+ \sum_{i=1,i\neq t}^{C} p^{\mathcal{T}}_{\backslash t} \hat p^{\mathcal{T}}_{i} \log(\frac{p^{\mathcal{T}}_{\backslash t}}{p^{\mathcal{S}}_{\backslash t}}).
	\end{split}
	\label{sup_reform_kd2}
\end{equation}
\end{small}
Since $p^{\mathcal{T}}_{\backslash t}$ and $p^{\mathcal{S}}_{\backslash t}$ are irrelevant to the class index $i$, we have:
\begin{equation}
	\begin{split}
	\sum_{i=1,i\neq t}^{C} p^{\mathcal{T}}_{\backslash t} \hat p^{\mathcal{T}}_{i} \log(\frac{p^{\mathcal{T}}_{\backslash t}}{p^{\mathcal{S}}_{\backslash t}}) &= p^{\mathcal{T}}_{\backslash t} \log(\frac{p^{\mathcal{T}}_{\backslash t}}{p^{\mathcal{S}}_{\backslash t}}) \sum_{i=1,i\neq t}^{C} \hat p^{\mathcal{T}}_{i} \\
	&= p^{\mathcal{T}}_{\backslash t} \log(\frac{p^{\mathcal{T}}_{\backslash t}}{p^{\mathcal{S}}_{\backslash t}}).
	\end{split}
\end{equation}
Then, 
\vspace{-1pt}
\begin{small}
\begin{equation}
	\begin{split}
	\text{KD} &= \underbrace{p^{\mathcal{T}}_{t}\log(\frac{p^{\mathcal{T}}_{t}}{p^{\mathcal{S}}_{t}}) + p^{\mathcal{T}}_{\backslash t} \log(\frac{p^{\mathcal{T}}_{\backslash t}}{p^{\mathcal{S}}_{\backslash t}})}_{{\text{KL}(\mathbf{b}^{\mathcal{T}}||\mathbf{b}^{\mathcal{S}})}} + p^{\mathcal{T}}_{\backslash t} \underbrace{\sum_{i=1,i\neq t}^{C} \hat p^{\mathcal{T}}_{i} \log(\frac{\hat p^{\mathcal{T}}_{i}}{\hat p^{\mathcal{S}}_{i}})}_{\text{KL}(\hat{\mathbf{p}}^{\mathcal{T}} || \hat{\mathbf{p}}^{\mathcal{S}})}.
	\end{split}
	\label{sup_reform_kd1}
\end{equation}
\end{small}
According to the definition of KL-Divergence, Eqn.(\ref{sup_reform_kd1}) can be rewritten as~(which is the same as Eqn.(\ref{form_dkd}) of the manuscript):
\begin{equation}
\begin{split}
%KD &= KL(\mathbf{P}^{b}_{\mathcal{T}}||\mathbf{P}^{b}_{\mathcal{S}}) + P^{\mathcal{T}}_{\backslash t} \sum_{i\neq t}^{C} \hat P^{\mathcal{T}}_{i} \log(\frac{\hat P^{\mathcal{T}}_{i}}{\hat P^{\mathcal{S}}_{i}})\\
\text{KD} &= \text{KL}(\mathbf{b}^{\mathcal{T}}||\mathbf{b}^{\mathcal{S}}) + (1-p_{t}^{\mathcal{T}}) \text{KL}(\hat{\mathbf{p}}^{\mathcal{T}} || \hat{\mathbf{p}}^{\mathcal{S}})
\end{split}
\label{sup_form_dkd}
\end{equation}

\subsection*{A.2. Implementation: Experiments in Sec~\ref{sec:exp}}
\label{sub:impl_details}

%For all experiments, we optimize the model with the commonly used loss setting~(\eg, cross-entropy for image classification) and the DKD loss.
\textit{\textbf{CIFAR-100}}: Our implementation for CIFAR-100 follows the practice in \cite{crd}. Teachers and students are trained for 240 epochs with SGD. As the batch size is 64, the learning rates are 0.01 for ShuffleNet\cite{shufflenetv2} and MobileNet-V2\cite{mobilenetv2}, 0.05 for the other series~(\eg VGG\cite{vgg}, ResNet\cite{resnet} and WRN\cite{wrn}). The learning rate is divided by 10 at 150, 180 and 210 epochs. The weight decay and the momentum are set to 5e-4 and 0.9. The weight for the cross-entropy loss is set to 1.0. The temperature is set as 4 and $\alpha$ is set as 1.0 for all experiments. The proper value of $\beta$ could be different for different teachers, and the details and discussions are in the next section. And we utilize a 20-epoch linear warmup for all experiments since the value of $\beta$ could be high leading to a large initial loss.  %done:说一下为什么放到supp里面？

\textit{\textbf{ImageNet}}: Our implementation for ImageNet follows the standard practice. We train the models for 100 epochs. As the batch size is 512, the learning rate is initialized to 0.2 and divided by 10 for every 30 epochs. Weight decay is 1e-4 and the weight for the cross-entropy loss is set to 1.0. We set temperature as 1 and $\alpha$ as 0.5 for all experiments. Strictly following~\cite{crd, revkd}, for distilling networks of the same architecture, the teacher is ResNet-34 model, the student is ResNet-18, and $\beta$ is set to 0.5. For different series, the teacher is ResNet-50 model, the student is MobileNet-V1, and $\beta$ is set to 2.0.

\textit{\textbf{MS-COCO}}: Our implementation for MS-COCO follows the settings in \cite{revkd}. We use the two-stage method Faster R-CNN\cite{faster_rcnn} with FPN\cite{fpn} as the feature extractors. ResNet\cite{resnet} models and MobileNet-V2\cite{mobilenetv2} are selected as teachers and students. All students are trained with the 1x scheduler~(schedulers and task-specific loss weights follow Detectron2\cite{detectron2}). We employ the DKD loss on the R-CNN head, and set $\alpha$ as 1.0, $\beta$ as 0.25, and temperature as 1 for all experiments.

Results of compared methods are reported in their original papers or reproduced by previous works~\cite{crd,revkd}.

\subsection*{A.3. Guidance for tuning $\beta$}
\label{sec_supp_a2}

We suppose that the importance of NCKD in knowledge transfer could be related to the confidence of the teacher. Intuitively, the more confident the teacher is, the more valuable the NCKD could be, and the larger $\beta$ should be applied. However, NCKD could increase the gradient contributed by logits of non-target classes. Thus, an improper large $\beta$ could harm the correctness of the student's prediction. If the logit value of the target class is much higher than all non-target classes, the teacher could be regarded as more confident and a large $beta$ could be more reasonable. Thus, we suppose that the value of $\beta$ could be related to the logit value \textit{gap} between the target and all non-target classes. Specifically, the \textit{gap} between the logit of the target class~(\ie, $z_{t}$, where $z$ represents the output logit and $t$ represents the target class) and the max logit among non-target classes could be reliable guidance for tuning $\beta$, which can be denoted as $z_t - z_{max}$, where $z_{max} = \text{max}(\{ z_i | i \neq t \}))$.

\begin{table}[h]
\setlength{\belowcaptionskip}{-10pt} %endtabspace
\center
\begin{small}
	\begin{tabular}{c|ccc}
	$\beta$ & ResNet56 & WRN-40-2 & ResNet32x4 \\ 
% 	Top-1 & 72.34 & 75.61 & 79.42 \\
	\Xhline{3\arrayrulewidth}
	1.0          & 76.02        & 75.94        & 74.95          \\
	2.0          & \textbf{76.32}        & 76.25        & 75.64          \\
	4.0          & 75.91        & 76.17        & 75.82          \\
	6.0          & 75.62        & \textbf{76.70}        & 76.34          \\
	8.0          & 75.33        & 76.44        & \textbf{76.45}          \\
	10.0         & 75.35        & 76.21        & 76.32       \\
	\hline
	$z_{t} - z_{max}$ & 5.36 & 7.24 & 8.40 \\
	\end{tabular}
\end{small}
\vspace{-8pt} %tablespace
\caption{Accuracy(\%) on CIFAR-100\cite{cifar} with different $\beta$ and different teachers. The \textit{gap}~($z_t - z_{max}$) is also reported.}
\label{tab:beta_pt}
\end{table}

We report experimental results on CIFAR-100\cite{cifar} to verify this conjecture. We select ResNet56, WRN-40-2 and ResNet32$\times$4 as teachers and ShuffleNet-V1 as the student, and apply different $\beta$. Both top-1 accuracy~(\%) and the \textit{gap} $z_{t} - z_{max}$~(averaged over all training samples) are reported. As shown in Table~\ref{tab:beta_pt}, the best value of $\beta$ could be positively proportional to the \textit{gap}, which we suppose could be guidance of tuning $\beta$ and a direction for further research. Based on this, the value of $\beta$ for each teacher in Table~\ref{tab:xifa} and Table~\ref{tab:xifa2} of the manuscript is set as follows~(in Table~\ref{tab:beta_for_cifar}):
\begin{table}[h]
\center
\begin{small}
\begin{tabular}{c|cc}
    teacher & $z_{t} - z_{max}$ & $\beta$ \\
    \Xhline{3\arrayrulewidth}
    ResNet56 & 5.36 & 2.0 \\
    ResNet110 & 6.73 & 2.0 \\
    WRN-40-2 & 7.24 & 6.0 \\
    VGG13 & 8.25 & 6.0 \\
    ResNet50 & 8.53 & 8.0 \\
    ResNet32$\times$4 & 8.40 & 8.0\\
    
\end{tabular}
\end{small}
\caption{The value of $\beta$ for different teachers in Table~\ref{tab:xifa} and Table~\ref{tab:xifa2} of the manuscript.}
\label{tab:beta_for_cifar}
\end{table}

\subsection*{A.4. Implementation: Experiments in Sec~\ref{sec:findings}}

In this part, we report the implementation details of the experiments in Sec~\ref{sec:findings} of the manuscript.

\vspace{5pt}
\noindent \textbf{Basic settings.}
We set the loss term of KD and CE as 1.0 and 1.0, respectively~(instead of the default $0.1\textit{CE}+0.9\textit{KD}$ setting in \cite{crd}). The setting in \cite{crd} follows the loss form proposed by \cite{kd}, which assumes that the sum of all terms' weights should be 1.0. However, the NCKD loss is target-irrelevant, which means the target-relevant loss could be 0.1 if we utilize the original setting when only applying NCKD. Based on this, we set the loss weight of all terms~(\eg, KD, TCKD, NCKD and CE) as 1.0 for all experiments in Sec~\ref{sec:findings} of the manuscript.

\vspace{5pt}
\noindent \textbf{Strong augmentation.}
We employ the AutoAugment\cite{autoaug} to reveal the effectiveness of TCKD in Sec~\ref{sec:findings} of the manuscript. Specifically, we add the CIFAR AutoAugment policy\footnote{\href{https://github.com/DeepVoltaire/AutoAugment}{https://github.com/DeepVoltaire/AutoAugment}} after applying the default augmentation~(random crop and horizontal flip). Then we train the teacher and the student with the same augmentation policy.

\vspace{5pt}
\noindent \textbf{Noisy labels.}
We also perform experiments on noisy training data to verify that TCKD conveys the knowledge about sample ``difficulty''. Specifically, we follow the settings of \cite{nosiy_label2,coteach}, utilizing the symmetric noise type\footnote{\href{https://github.com/bhanML/Co-teaching/blob/master/data/cifar.py}{https://github.com/bhanML/Co-teaching/blob/master/data/cifar.py}}. We train a teacher network on the noisy training data and select the best epoch to distill the student~(on the same training data).

\subsection*{A.5. Explanation about why TCKD brings performance drop in Table~\ref{tab:tckd_nckd}}
In Table~\ref{tab:tckd_nckd} of the manuscript, we reveal that singly applying TCKD could bring performance drop sometimes. An explanation for this phenomenon is that the high temperature~(T=4) will lead to a great gradient to increase the non-target classes' logits, which will harm the correctness of the student's prediction. Without NCKD, the information about the class similarity~(or the prominent dark knowledge) is not available, so that TCKD's gradient could do no good but lead to performance drop~(since TCKD could bring marginal performance gain on easy-fitting training data). To verify that the large temperature is not proper when singly applying TCKD, we perform experiments with different temperatures~(T) in the table below. 
\begin{table}[h]
\center
\begin{small}
\begin{tabular}{c|cccc}
    T & 1 & 2 & 3 & 4 \\
    \Xhline{3\arrayrulewidth}
    top-1 & 73.24 & 73.05 & 71.69 & 70.96 \\
    
\end{tabular}
\end{small}
\caption{Accuracy~(\%) with different temperature(T) when only applying TCKD. The teacher and the student are set as WRN-40-2 and WRN-16-2, respectively.}
\label{tab:diff_T_tckd}
\end{table}
Results in Table~\ref{tab:diff_T_tckd} show that the performance is almost the same as the vanilla training baseline~(73.26 in Table~\ref{tab:tckd_nckd} of the manuscript) when the temperature is set as 1. And the performance drop is positively related to the temperature.

\subsection*{A.6. How to employ DKD on detectors}
In this paper, we employ our DKD on the two-stage object detector Faster R-CNN.
We only employ our DKD on the R-CNN head. Specifically, given a student network, we utilize the labels assigned to the proposals~(generated by the RPN module) as the target class(if $\textit{IoU(proposal)}<0.5$, the target class is set as ``background''). Then, we use a teacher network to get the R-CNN prediction logits of the \textit{same proposals}~(locations are the same, while the features are from the teacher's backbone). Thus, we can employ our DKD by minimizing the KL-Divergence~(\ie, TCKD and NCKD) between the student's logits and the teacher's.

\subsection*{A.7. Implementation: Experiments in Sec~\ref{sec:ext}}

\noindent \textbf{Training efficiency.}
We report the training time of each distillation method in Figure~\ref{fig:speed} of the manuscript. The training time~(per batch) is the sum of (1) the data processing time~(\eg, including the time to sample the contrast examples in \cite{crd}), (2) the network forward time and the gradient backward time and (3) the memory updating time~(\eg,including the time to update the contrast memory in \cite{crd}).
We also report the number of extra parameters for each method. Besides the learnable parameters~(\eg, the connectors in \cite{ofd} and the ABF modules in \cite{revkd}), we also calculate the extra dictionary memory(\eg, the contrast memory in \cite{crd}).

\vspace{5pt}
\noindent \textbf{Feature transferability.}
We perform linear probing experiments to verify the feature transferability of our DKD in Sec~\ref{sec:ext} of the manuscript. We use the WRN-16-2 distilled from a WRN-40-2 teacher as the feature extractor~(only using the feature generated by the final global average pooling module), then train linear fully-connected~(FC) layers as classifier modules for STL-10 and Tiny-ImageNet datasets~(the feature extractor is fixed during training). We train the FC via an SGD optimizer with 0.9 momentum and 0.0 weight decay. The number of total epochs is set as 40, and the learning rate is set to 0.1 for a 128 batch size and divided by 10 for every 10 epochs.

% 速度-性能trade-off图
% \clearpage

% \begin{table}[h]
% \setlength{\belowcaptionskip}{-10pt} %endtabspace
% \begin{small}
% \center
% 	\begin{tabular}{c|ccc}
% 	& top-1 & time & params \\ \Xhline{3\arrayrulewidth}
% 	KD\cite{kd}          & 73.33        &  11       & 0          \\
% 	RKD\cite{rkd}          & 71.90        & 25        & 0          \\
% 	FitNet\cite{fitnets}          & 73.50        & 14        & 16.8K          \\
% 	OFD\cite{ofd}          & 74.95        & 19        & 86.9K          \\
% 	CRD\cite{crd}         & 75.51        & 41        & 12.3M \\
% 	ReviewKD\cite{revkd}    & 75.63        & 26        & 1.8M \\ \hline
% 	\textbf{DKD}         & \textbf{76.32}        & \textbf{11}        & \textbf{0}        
% 	\end{tabular}
% \end{small}
% \end{table}

\end{document}